\documentclass[11pt,a4paper]{article}
\usepackage[T1]{fontenc}
\usepackage[utf8]{inputenc}
\usepackage{lmodern}
\usepackage[margin=25mm]{geometry}
\usepackage{microtype}
\usepackage{graphicx}
\usepackage{booktabs}
\usepackage{amsmath,amssymb}
\usepackage{tabularx}
\usepackage[numbers,square,sort&compress]{natbib}
\usepackage[hyphens]{url}
\usepackage{hyperref}
\hypersetup{hidelinks,
  pdftitle={CompositionEvidence: Controlling Geometric Evidence for Cinematic Composition Reasoning with Frozen Vision-Language Models},
  pdfauthor={Yichen He, Xin Huang, Yuxi Liu},
  pdfsubject={Cinematic composition reasoning with frozen vision-language models}}
\title{\textbf{CompositionEvidence: Controlling Geometric Evidence for Cinematic Composition Reasoning with Frozen Vision--Language Models}}
\author{\parbox{\dimexpr\textwidth-2\tabcolsep\relax}{\centering
  Yichen~He\textsuperscript{1}\quad
  Xin~Huang\textsuperscript{1}\quad
  Yuxi~Liu\textsuperscript{2}\thanks{Corresponding author: \href{mailto:26b951179@stu.hit.edu.cn}{26b951179@stu.hit.edu.cn}.}\\[8pt]
  {\small\textsuperscript{1}School of Journalism and Communication, Bohai University,\\
  Jinzhou, Liaoning, China}\\[3pt]
  {\small\textsuperscript{2}School of Computer Science and Technology,\\
  Harbin Institute of Technology (Shenzhen), Shenzhen, Guangdong, China}\\[7pt]
  {\small\href{mailto:heyichen626@gmail.com}{heyichen626@gmail.com}\quad
  \href{mailto:huangxin@qymail.bhu.edu.cn}{huangxin@qymail.bhu.edu.cn}\\
  \href{mailto:26b951179@stu.hit.edu.cn}{26b951179@stu.hit.edu.cn}}\\[7pt]
  {\footnotesize ORCID: Yichen He~\href{https://orcid.org/0009-0006-1932-4461}{0009-0006-1932-4461}\\
  Xin Huang~\href{https://orcid.org/0009-0008-7078-8733}{0009-0008-7078-8733}\quad
  Yuxi Liu~\href{https://orcid.org/0009-0006-3855-6735}{0009-0006-3855-6735}}
}}
\date{}
\begin{document}
\maketitle

\begin{abstract}
Cinematic composition questions depend on in-frame geometry that
vision--language models (VLMs) use inconsistently. External perception tools can
expose this geometry, but naive prompt injection can overwrite useful visual
predictions. On ShotBench composition, pre-question geometry reduces
Qwen2.5-VL-7B from 49.5\% to 46.3\%; corrupted measurements lower it to 33.6\%.
\emph{CompositionEvidence} derives nine reproducible features from frozen
detection, depth, and line experts and supplies their shared 24-dimensional
representation through two interfaces. A training-free channel appends gated
neutral evidence after the question, reaching 56.2\% ($+6.7$, exact McNemar
$p{=}0.0018$) and 48.4\% under corrupted evidence. A minimal-training channel
maps the same representation to four soft tokens with a 7.37M-parameter
projector while leaving the VLM frozen; it reaches 61.0\% ($+11.5$). Across nine
VLMs, including five commercial models queried through official provider APIs,
zero-shot accuracy ranges from 43.8\% to 63.0\%, and the text channel changes it
by $+6.7$ to $-6.9$ points; the highest baseline is among the regressions.
The effect of explicit geometry depends on measurement reliability and on how the
receiving model integrates it.

\par\medskip\noindent\textbf{Keywords:} Cinematic composition, Vision--language models, Geometric
evidence, Tool-augmented reasoning, Frozen backbones
\end{abstract}

\section{Introduction}
Cinematic composition carries meaning through geometry, such as where a subject
sits, how visual mass is distributed across the frame, and whether a shot is
mirror-alike. Current frontier VLMs are broadly capable, but they use this kind of
fine-grained evidence unevenly. In our nine-model
ShotBench~\cite{liu2025shotbench} evaluation, zero-shot accuracy spans
43.8\%--63.0\%, and the same geometric evidence produces gains on some models and
regressions on others. This paper studies how the evidence is delivered to a VLM,
and every controlled comparison keeps the receiving model fixed.

Composition questions require information that general-purpose vision encoders are not trained to preserve. Such encoders are optimised for semantic alignment, while composition decisions turn on small geometric quantities such as the offset of a centroid from a reference line, the difference in left/right visual mass, mirror similarity, and depth layering. Frozen detection, monocular-depth, and line-segment experts can measure these quantities and pass them to the VLM.

Supplying the measurements does not by itself improve the model. When we place the geometry before the question, Qwen2.5-VL-7B drops from 49.5\% to 46.3\% and 165 of 479 answers change. Replacing the measurements with internally consistent but incorrect values lowers accuracy to 33.6\%, which is above the 25\% chance rate and 15.9 points below zero-shot. The injected block displaces part of the model's visual prior, so its position and its reliability need as much attention as the extraction procedure itself.

\emph{CompositionEvidence}\footnote{Research artifact (code, per-sample records, supplementary material): \url{https://github.com/Cyber-Yichen/composition-evidence-artifact}.} supplies the same nine formalised geometric features through two interfaces. Channel A is training-free and compatible with closed APIs. It lets the VLM read the image, question, and options before appending a controlled evidence block whose uncertain lines are suppressed by training-fitted thresholds. This ordering and filtering raise ShotBench composition accuracy to 56.2\% ($+6.7$, $p{=}0.0018$). Channel B uses a two-layer projector to map the 24-dimensional feature vector to four soft tokens while the VLM remains frozen. Training only 7.37M parameters yields 61.0\% ($+11.5$), outperforming the published ShotVL-3B score of 57.4\% and coming within 1.2 points of fully fine-tuned ShotVL-7B v1.1. Both channels reuse identical cached features, which isolates the interface choice.

We make three contributions.
\begin{itemize}
    \item We formalise nine composition features computed from frozen detection, depth, and line experts, and deliver the resulting 24-dimensional vector through two interfaces, confidence-gated post-question text and a learned four-token projection.
    \item We isolate the effect of the interface with paired transition counts and corrupted-evidence controls, and show how pre-question injection replaces correct baseline decisions.
    \item We measure where each interface helps and where it hurts, using interface ablations, the seven non-composition ShotBench dimensions, a second benchmark, and nine VLMs.
\end{itemize}

\section{Related Work}
\paragraph{Cinematic understanding.}
Early computational film analysis treated shot scale and camera motion as separate classification tasks~\cite{rao2020movieshots,huang2020movienet}. ShotBench~\cite{liu2025shotbench} unifies cinematic understanding into eight dimensions with 3{,}608 expert QA pairs and introduces ShotQA and the ShotVL specialists. ShotVL's evidence-free chain-of-thought SFT offers little improvement over plain SFT. RefineShot~\cite{wu2025refineshot} restructures parts of ShotBench, but its composition subset is identical to the original. We therefore use CineTechBench~\cite{wang2025cinetechbench}, which has a different label taxonomy, for cross-benchmark evaluation.

\paragraph{Structured evidence for multimodal reasoning.}
CamReasoner~\cite{wu2026camreasoner} combines structured camera-motion reasoning with reinforcement learning, while Geometry-Guided Camera Motion~\cite{feng2026geometryguided} injects 3D foundation cues into a frozen VideoLLM, and explicit spatial cues have also been shown to improve multimodal spatial reasoning~\cite{zhang2026camcue}. We study how a VLM responds to imperfect external measurements, and our paired and corrupted-evidence controls separate the effect of the injection interface from the information the perception tools supply.

\paragraph{Spatial reasoning and external tools.}
Diagnostic datasets from CLEVR~\cite{johnson2017clevr} and GQA~\cite{hudson2019gqa} to SpatialSense~\cite{yang2019spatialsense}, VSR~\cite{liu2023vsr}, and What'sUp~\cite{kamath2023whatsup} repeatedly expose a gap between semantic recognition and relational geometry. SpatialVLM narrows it through large-scale spatial supervision~\cite{chen2024spatialvlm}; we instead retain a general frozen VLM and supply compact, auditable measurements at inference. Tool-augmented systems such as VisProg and ViperGPT compose specialist vision modules through generated programs~\cite{gupta2023visprog,suris2023vipergpt}, and Toolformer learns when to invoke textual APIs~\cite{schick2023toolformer}. We study the failure mode that follows. Even when a tool has returned a useful value, the receiving model may over-trust it.

\paragraph{Efficient conditioning and calibration.}
Frozen-backbone interfaces commonly learn a bridge from visual features to language-model inputs~\cite{alayrac2022flamingo,li2023blip2}; prompt and prefix tuning adapt frozen models with learned input vectors~\cite{lester2021prompt,li2021prefix,jia2022vpt}, and PIN shows that a small positional insert can unlock localisation in a frozen VLM~\cite{dorkenwald2024pin}. In Channel B the tokens encode a deterministic function of explicit composition measurements, whereas prompt tuning learns free task parameters. Channel A is closer to selective prediction~\cite{geifman2019selectivenet}, although it gates individual evidence statements and always answers. It also differs from probability calibration~\cite{guo2017calibration}, since what we calibrate is the \emph{influence} of an external evidence source.

\paragraph{Composition analysis and generation.}\leavevmode\\
Photographic composition guidance~\cite{you2026photoframer}, composition-preserving aesthetic assessment~\cite{mai2016aesthetics}, and aesthetic cropping~\cite{wei2018goodview,li2020composing} support shooting or view recommendation, and generation-side control asks how to produce a layout. We ask whether a language model can interpret the composition of an existing cinematic frame from explicit geometric evidence.

\section{CompositionEvidence}\label{sec:method}
\subsection{Task and Interfaces}
\begin{figure}[tb]
\centering
\includegraphics[width=\linewidth]{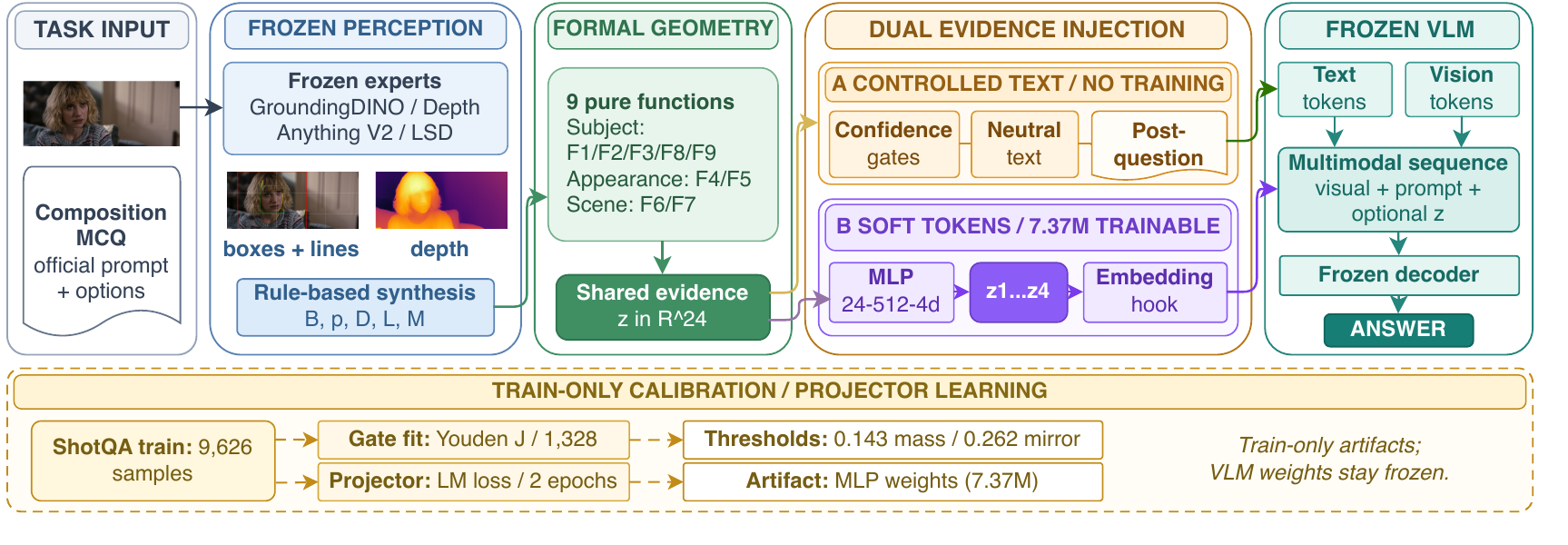}
\caption{\emph{CompositionEvidence} architecture. Frozen detection, depth, and line
experts feed deterministic synthesis and nine geometric functions, which yield the
shared vector $\mathbf z$. Channel A gates and appends selected measurements as
post-question text; Channel B maps $\mathbf z$ into four evidence tokens. Only the
fitted thresholds and the projector reach inference.}
\label{fig:architecture}
\end{figure}

Given a film still $I$ and a multiple-choice composition question $q$, we seek
to improve the VLM prediction without changing its weights (Channel A), or by
training only a small adapter (Channel B). Figure~\ref{fig:architecture} traces
both the inference path and the train-only construction of its artifacts. The
official image, question, and options remain the VLM task input. In parallel,
GroundingDINO, Depth Anything V2, and LSD produce boxes, depth, and line
segments; deterministic synthesis collects them as $B,p,D,L,M$. Nine pure
functions (five subject, two appearance, and two scene features) then form the
shared vector $\mathbf z\in\mathbb{R}^{24}$ used by both channels.

Channel A exposes only the measurements that pass their confidence gates, renders
them without answer-label language, and places the block after the MCQ before
tokenisation. Channel B maps the same $\mathbf z$ to four soft tokens that replace
four placeholder embeddings, leaving the vision encoder, tokeniser, and decoder
unchanged. The lower strip of Figure~\ref{fig:architecture} runs only at training
time. There, 1{,}328 of the 9{,}626 ShotQA samples fit the mass and mirror
thresholds with Youden $J$, and the full composition pool supplies the two-epoch
language-model loss that learns the projector. Only the fitted thresholds and the
projector weights cross into inference.

\subsection{Evidence Representation}
Let $\phi(I)$ denote the frozen perception-and-synthesis stack and
$\mathbf z=\phi(I)\in\mathbb{R}^{24}$ its continuous values and presence flags.
For Channel A, renderer $r_j$ verbalises feature group $j$, while a gate
$g_j(\mathbf z;\tau_j)\in\{0,1\}$ decides whether that line is exposed:
\begin{equation}
E_A(\mathbf z)=\mathop{\Vert}_{j=1}^{9}
  \left[g_j(\mathbf z;\tau_j)\,r_j(\mathbf z)\right],
\qquad
\hat y_A=f_\theta(I,q,E_A).
\end{equation}
The concatenation omits a term when its gate is zero. Thresholds $\tau_j$ and
percentile annotations are functions only of ShotQA training statistics. For
Channel B, a projector $P_\psi$ produces four evidence embeddings,
\begin{align}
\mathbf h&=\operatorname{GELU}(W_1\mathbf z+b_1),\\
T_B(\mathbf z)&=\operatorname{reshape}_{4\times d}(W_2\mathbf h+b_2),
\quad \hat y_B=f_\theta(I,q,T_B),
\end{align}
where $\theta$ is frozen and only $\psi=\{W_1,b_1,W_2,b_2\}$ is learned. Both
channels hold the evidence source and VLM fixed; they differ in how the
measurements enter the model. All controlled comparisons reuse the same cached
features, so the ablations compare interfaces without changing the detector or
base VLM. Channel A updates no parameters and remains compatible with API models,
whereas Channel B trains only $\psi$ on ShotQA. No test image changes the feature
definitions.

\subsection{Geometric Evidence Extraction}
The perception layer contains three frozen experts and one deterministic synthesis module. GroundingDINO~\cite{liu2024groundingdino}, queried first for people and then with a generic-subject fallback, produces boxes; Depth Anything V2~\cite{yang2024depthanythingv2} estimates relative depth; and the OpenCV LSD detector~\cite{vongioi2010lsd} extracts line segments. We synthesise pseudo-saliency from confidence-weighted box masks and depth closeness. A fixed rule selects the dominant subject from area, centre prior, and confidence, weighting person classes by $1.2\times$ and discarding any box that covers more than 90\% of the frame.

Table~\ref{tab:features} lists the nine features computed from these outputs. Each feature is a pure function of the synthesised observations and is covered by synthetic-image unit tests. F4 uses SSIM~\cite{wang2004ssim} between a frame and its mirror image. F6 uses equal-width depth bins; an earlier quantile design produced layer shares that were constant by construction.

\begin{table}[tb]
\centering
\caption{The nine geometric features and the Channel-A gates. Channel B always
receives the corresponding continuous values and the presence flags.}
\label{tab:features}
\footnotesize
\setlength{\tabcolsep}{4pt}
\begin{tabular}{@{}ll>{\raggedright\arraybackslash}p{0.40\linewidth}@{}}
\toprule
ID & Quantity & Channel A exposes it when \\
\midrule
F1 & Subject centroid $(x,y)$ & a dominant subject is detected \\
F2 & Distance to midpoint and thirds & F1 is exposed \\
F3 & Subject area ratio & F1 is exposed \\
F4 & Mirror similarity (SSIM) & horizontal SSIM outside $[0.144,0.262]$ \\
F5 & Left/right visual mass $b\in[-1,1]$ & $|b|\geq0.143$ \\
F6 & Relative depth layers & a depth map is available \\
F7 & Leading lines and horizon & line extraction succeeds \\
F8 & Head space & a face is detected \\
F9 & Edge proximity and full span & within 0.05 of an edge, or spanning a side \\
\bottomrule
\end{tabular}
\end{table}

\subsection{Channel A: Post-Question Gated Text}
Channel A renders measurements rather than answers. Any expression sharing a word root with the benchmark's label vocabulary is forbidden (for ShotBench: \emph{center/centre}, \emph{symmetr-}, \emph{balanc-}, \emph{heavy}, \emph{short side}), and generation stops if the renderer detects a violation. On a benchmark with different labels, the ban is re-derived from that taxonomy. We also add training-distribution percentiles to quantities such as mirror SSIM and mass difference, because their values on film frames are far below intuitive natural-image scales.

Two choices calibrate how the VLM uses this text. First, we append the block
\emph{after} the image question, options, and answering instruction, so the model
forms a task-specific prior before it reads the external measurements. A similar
ordering sensitivity has been reported for long text contexts in language
models~\cite{liu2024lost}; here it appears in a short multimodal prompt, and we
isolate it with paired and corrupted-evidence controls. Second, we expose a
measurement only when its signal is reliable, using the gates in
Table~\ref{tab:features}. Neutral facts such as subject position, depth, and lines
are always exposed. We fit the numeric thresholds on ShotQA training data with the
Youden $J$ statistic and keep them fixed for every reported run. We added the
gating mechanism after a first pass of ShotBench diagnostics, and
Section~\ref{sec:validity} discusses this design-selection effect.

\subsection{Channel B: Soft-Token Projection}
Channel B uses the same evidence without textual rendering. A two-layer MLP maps the 24 continuous features and presence flags through $24{\rightarrow}512{\rightarrow}4{\times}d$, where $d{=}3{,}584$ for Qwen2.5-VL-7B, and the four outputs replace placeholder embeddings through an embedding-layer hook. The VLM stays frozen; the projector contains $(24{\times}512+512)+(512{\times}4d+4d)=7{,}367{,}168$ trainable parameters. A seed-42 shuffle splits the 9{,}626 ShotQA composition samples into 9{,}145/481 train/validation items. We train two epochs with AdamW (learning rate $10^{-4}$, weight decay 0.01), effective batch size 8, gradient clipping at 1.0, and cosine scheduling; the projector is FP32, the frozen VLM BF16, and validation loss selects the checkpoint. Training takes a few hours on one RTX 4090D.

\section{Experiments}\label{sec:exp}
Our evaluation asks three questions. \textbf{RQ1:} Can calibrated geometry improve composition reasoning without updating a VLM, and with only minimal training? \textbf{RQ2:} Why does naive injection fail, and which design choices repair it? \textbf{RQ3:} Is the effect specific to one model or benchmark?

\subsection{Setup}\label{sec:setup}
We evaluate all 479 questions in the ShotBench composition subset, using its official four-option prompt. The labels are \emph{Center}, \emph{Symmetrical}, \emph{Balanced}, \emph{Left heavy}, \emph{Right heavy}, \emph{Short side}, and a few combinations. Our Qwen2.5-VL-7B~\cite{bai2025qwen25vl} zero-shot reproduction is 49.5\%, within two correct answers of the official 49.9\%. Inference uses greedy decoding and a fixed seed; predictions are logged per sample and compared with paired exact McNemar tests~\cite{mcnemar1947}.

The gates are calibrated on 1{,}328 ShotQA training images (at most 300 per class). Calibration moves the mirror-SSIM threshold from the intuitive 0.7 to 0.262 and the visual-mass threshold from 0.25 to 0.143, and puts pseudo-saliency direction precision at only 67.8\%, which motivates selective rendering. Besides the primary model we test InternVL3-8B~\cite{zhu2025internvl3}, LLaVA-OneVision-7B~\cite{li2024llavaonevision}, ShotVL-7B v1.1~\cite{liu2025shotbench}, and five commercial VLMs queried through their official provider APIs in July 2026.

\paragraph{Determinism and paired evaluation.}
Every local condition evaluates all 479 test items once with greedy decoding, seed
42, and at most 64 generated tokens, using Python 3.12.3, PyTorch 2.12.1,
Transformers 4.57.1, and one 24GB GPU. All systems answer the same items, so we
report exact McNemar tests on wrong-to-right versus right-to-wrong transitions
instead of treating the accuracies as independent samples. We retain per-sample
predictions, features, rendered blocks, thresholds, and checkpoint hashes, so every
aggregate in the paper can be reconstructed.

\subsection{RQ1: Results on Qwen2.5-VL-7B}
\begin{table}[tb]
\centering
\caption{ShotBench composition results with Qwen2.5-VL-7B ($n{=}479$). Both
calibrated channels improve the same zero-shot model; reasoning or raw evidence
alone does not. ShotBench reference scores: GPT-4o 55.2, ShotVL-3B 57.4,
fine-tuned ShotVL-7B v1.1 62.2.}
\label{tab:main}
\setlength{\tabcolsep}{3pt}
\footnotesize
\begin{tabular}{lccl}
\toprule
Condition & Acc.\ (\%) & $\Delta$ vs.\ C0 & Exact McNemar \\
\midrule
Rule classifier (no VLM) & 37.8 & $-11.7$ & --- \\
C0 zero-shot (official prompt) & 49.5 & --- & --- \\
Chain-of-thought, no evidence & 47.4 & $-2.1$ & --- \\
Naive evidence (pre-question) & 46.3 & $-3.1$ & $p{=}0.276$ (75$\uparrow$/90$\downarrow$) \\
Post-question, ungated & 53.4 & $+3.9$ & --- \\
\textbf{Channel A: post-question + gates} & \textbf{56.2} & \textbf{+6.7} & $p{=}0.0018$ (66$\uparrow$/34$\downarrow$) \\
\textbf{Channel B: four soft tokens (7.37M)} & \textbf{61.0} & \textbf{+11.5} & $p{<}10^{-5}$ (100$\uparrow$/45$\downarrow$) \\
\bottomrule
\end{tabular}
\end{table}

Both calibrated channels improve the same zero-shot model, while reasoning or raw
evidence alone does not (Table~\ref{tab:main}). Channel A gains 6.7 points without
model-specific training. Channel B gains 11.5 points, exceeds ShotVL-3B (57.4\%),
and comes within 1.2 points of fully fine-tuned ShotVL-7B v1.1 while training
roughly three orders of magnitude fewer parameters. The rule classifier reaches
37.8\% and chain-of-thought without evidence falls below zero-shot, so the gains
require the VLM prediction and the added geometry together. Current commercial
baselines appear in Figure~\ref{fig:ninemodel}.

The paired counts in Table~\ref{tab:main} show which predictions change. Channel A
nets 32 corrected items and Channel B 55, while naive injection changes the most
answers (165) and breaks more correct predictions than it repairs.

\begin{figure}[tb]
\centering
\includegraphics[width=0.315\linewidth]{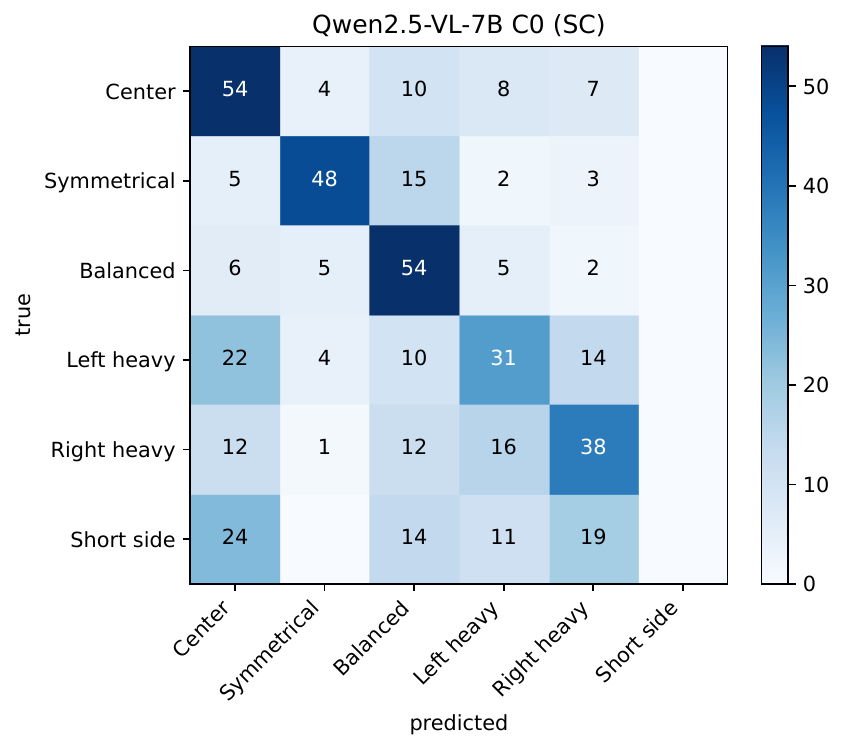}\hfill
\includegraphics[width=0.315\linewidth]{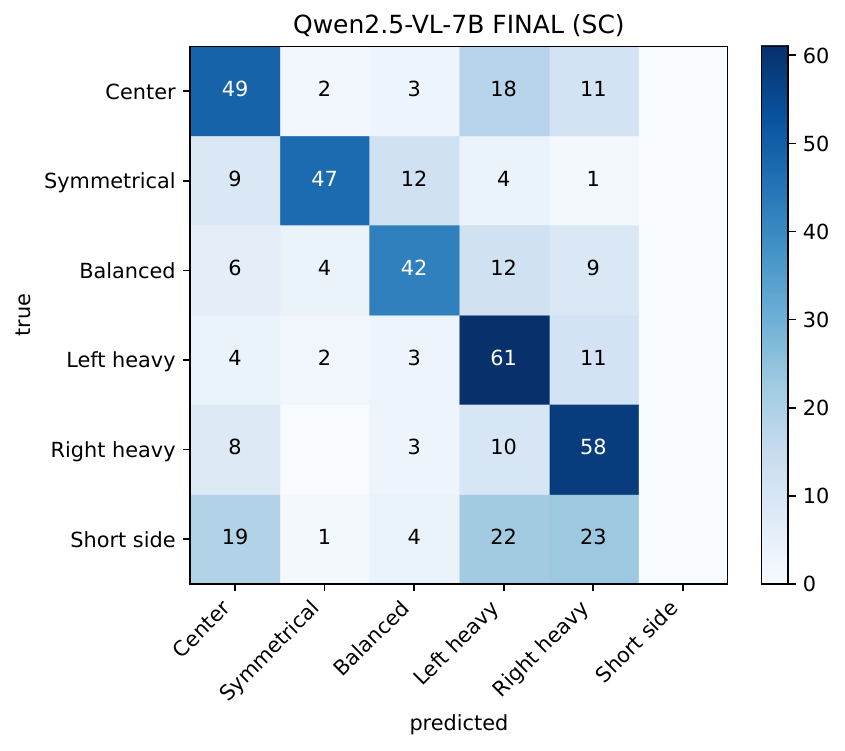}\hfill
\includegraphics[width=0.315\linewidth]{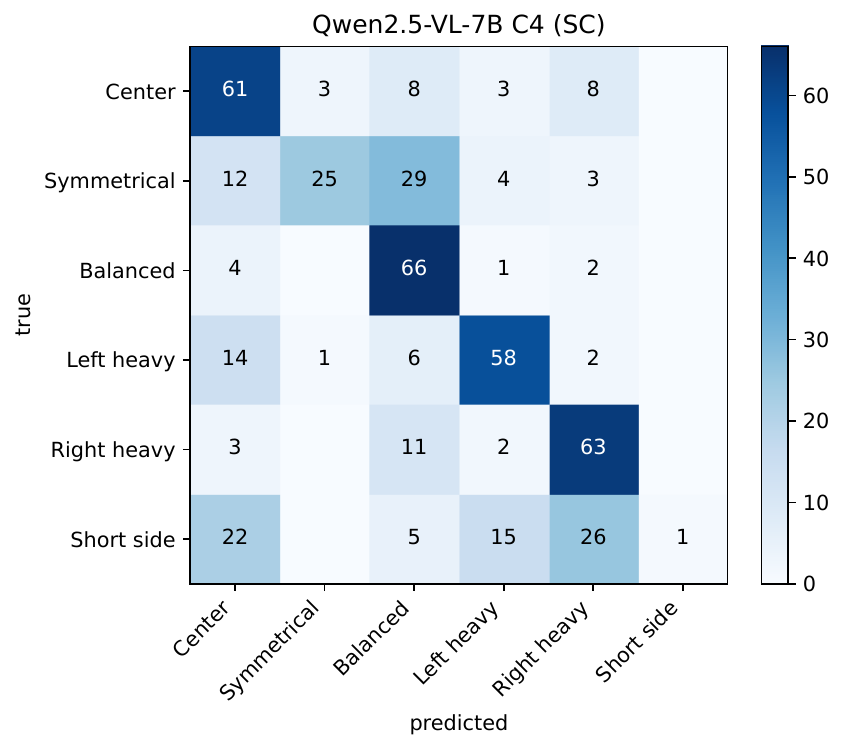}
\caption{Confusion matrices for zero-shot (left), Channel A (middle), and Channel B
(right). Calibrated evidence recovers directional classes while preserving much of
the baseline diagonal; \emph{Short side} remains nearly absent.}
\label{fig:confusion}
\end{figure}

The gains concentrate on the classes that geometry resolves directly. Under Channel A, \emph{Left heavy} rises from 38.3\% to 75.3\% and \emph{Right heavy} from 48.1\% to 73.4\%. \emph{Balanced} falls from 74.0\% to 57.5\%, which is the main residual cost of noisy visual-mass evidence, and \emph{Short side} stays at zero. Figure~\ref{fig:confusion} shows the redistribution across the full label space.

\subsection{RQ2: Interface Ablations and Corruption Controls}\label{sec:ablation}
\paragraph{Pre-question injection.}
Putting evidence before the question lowers accuracy to 46.3\%, but the aggregate understates the effect: 75 baseline errors are corrected while 90 correct predictions are broken. Numeric-only and neutral-language renderings behave similarly, ruling out wording as the main cause, and a single feature group (48.2\%) outperforms the complete pre-question block. The evidence changes the answer distribution even when its net effect is negative.

\paragraph{Post-question placement and gating.}
Moving the same ungated block after the question raises accuracy to 53.4\%, an increase of 7.1 points over pre-question injection. Confidence control then improves it to 54.5\% with strict precision gates and to 56.2\% with Youden gates. A $3{\times}3$ sweep over the centre/thirds alignment threshold ($0.05$--$0.12$) and the strong-symmetry threshold ($0.6$--$0.8$) stays within 56.8--57.0\%, so the conclusion does not depend on one threshold combination. Position protects the model's task prior, and gating limits the influence of unreliable measurements.

\begin{table}[tb]
\centering
\caption{Interface ablation with the evidence source and Qwen2.5-VL-7B fixed.
Wording alone does not repair pre-question injection; position supplies the
largest step, and selective coverage adds the remaining gain.}
\label{tab:interface}
\footnotesize
\setlength{\tabcolsep}{3.5pt}
\begin{tabular}{llc}
\toprule
Evidence interface & Control & Acc.\ (\%) \\
\midrule
Pre-question, full & none & 46.3 \\
Pre-question, numeric & none & 45.1 \\
Pre-question, one group & none & 48.2 \\
Post-question, full & none & 53.4 \\
Post-question, full & strict gates & 54.5 \\
Post-question, full & Youden gates & \textbf{56.2} \\
\bottomrule
\end{tabular}
\end{table}

\begin{figure}[tb]
\centering
\includegraphics[width=0.98\linewidth]{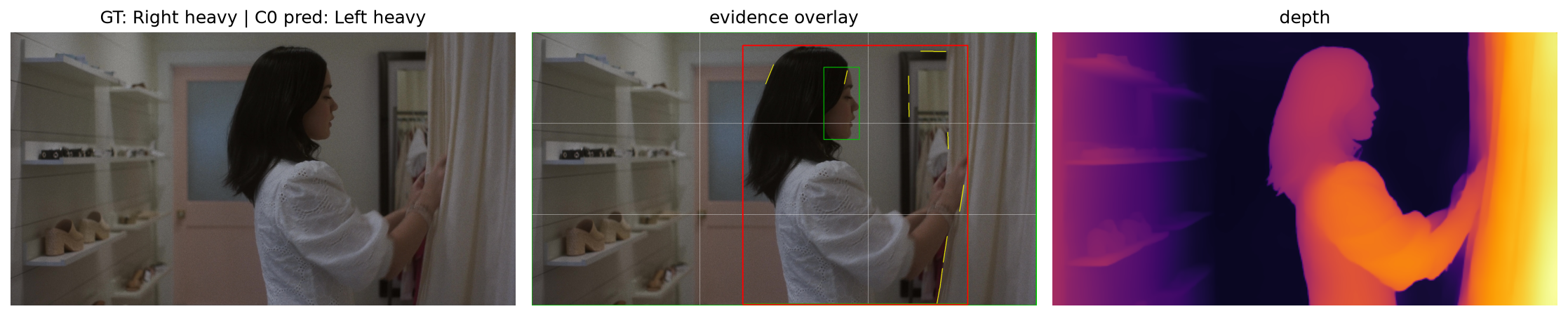}
\vspace{2pt}

\begin{minipage}{\linewidth}
\scriptsize
\begin{tabularx}{\linewidth}{>{\bfseries}p{0.16\linewidth}X}
Question & What's the composition of this shot?
\quad A Right heavy \quad B Symmetrical \quad C Left heavy \quad D Short side \\
Ground truth & A --- Right heavy \\
Before (C0) & C --- Left heavy. The baseline treats the empty space on the right as visual weight. \\
Measured evidence & Dominant-subject centroid $x{=}0.64$; left/right visual mass $0.36/0.64$;
difference $0.27$ (78th training percentile). \\
After A\,/\,B & A --- Right heavy \quad/\quad A --- Right heavy \\
What changed & Calibrated geometry distinguishes occupied visual mass from negative space, reversing the directional error. \\
\end{tabularx}
\end{minipage}
\caption{\textbf{A Before/After Q\&A demo} (\texttt{shotbench-2638}), drawn from the
pool of items that C0 answers incorrectly and Channel A answers correctly. The
supplementary material adds two further demos and a counterexample in which
Channel A breaks a correct answer.}
\label{fig:demo-main}
\end{figure}

\paragraph{Response to corrupted evidence.}
We replace the evidence with internally consistent but incorrect values by randomising the centroid, re-deriving the dependent quantities, flipping the mass direction, and randomising mirror similarity. Naive pre-question injection then falls to 33.6\%, which is 15.9 points below the no-evidence baseline, whereas Channel A yields 48.4\%. The calibrated interface therefore relies less on the block when its fields conflict with the image.

Component-wise corruption separates evidence quality from interface robustness.
Position, mass, and mirror corruption leave Channel A at 54.1\%, 51.1\%, and
56.4\%, so mass is both the most useful cue and the largest error source,
consistent with its 67.8\% precision and the remaining \emph{Balanced} errors. The
7.8-point clean-to-corrupt drop shows that the tools matter, and the one-point
distance to C0 shows that they no longer overwhelm the image.

\begin{table}[tb]
\centering
\caption{Clean utility and corrupted-evidence fallback are distinct. Naive
injection fails both tests; calibrated rendering gains on clean evidence and
returns close to the image-only prior when every field is wrong.}
\label{tab:corruption}
\footnotesize
\setlength{\tabcolsep}{4pt}
\begin{tabular}{lcc}
\toprule
Evidence condition & Acc.\ (\%) & $\Delta$ vs.\ C0 \\
\midrule
C0: no evidence & 49.5 & --- \\
Naive: clean & 46.3 & $-3.1$ \\
Naive: all fields corrupted & 33.6 & $-15.9$ \\
Channel A: clean & \textbf{56.2} & $\mathbf{+6.7}$ \\
\quad position corrupted & 54.1 & $+4.6$ \\
\quad visual mass corrupted & 51.1 & $+1.6$ \\
\quad mirror similarity corrupted & 56.4 & $+6.9$ \\
\quad all fields corrupted & 48.4 & $-1.0$ \\
\bottomrule
\end{tabular}
\end{table}

\paragraph{Clean accuracy and corrupted-evidence fallback.}
Let $A(\varnothing)$, $A(E)$, and $A(\widetilde E)$ denote image-only, measured, and
corrupted-evidence accuracy. The difference $A(E)-A(\varnothing)$ measures the
useful signal a channel extracts, and $A(\widetilde E)-A(\varnothing)$ measures how
far it falls back to the visual prior. Both quantities matter, because an interface can repair a few errors while
creating a much larger failure surface. Channel A moves the pair from
$(-3.1,-15.9)$ for naive injection to $(+6.7,-1.0)$ points. The image, question,
options, and decoder are fixed throughout, so the two differences isolate the
model's reliance on the injected fields.

\subsection{RQ3: Other Models and a Second Benchmark}
\begin{figure}[tb]
\centering
\includegraphics[width=\linewidth]{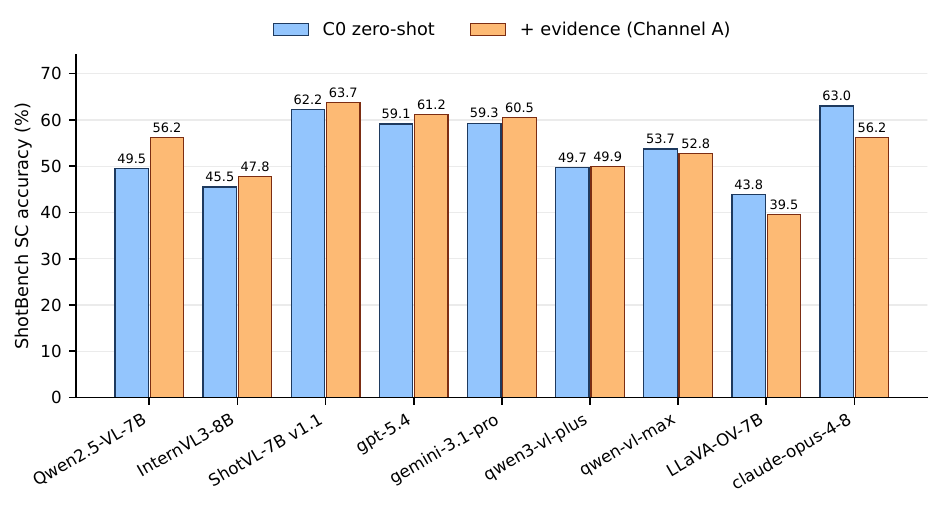}
\caption{The training-free recipe applied unchanged to nine VLMs on ShotBench
composition. The five commercial models are queried through official provider APIs.}
\label{fig:ninemodel}
\end{figure}

Five models improve, two stay approximately flat, and two decline
(Figure~\ref{fig:ninemodel}), and the gains are not monotone in baseline accuracy.
The primary model with Channel A (56.2\%)
stays below all three commercial baselines, and Channel B (61.0\%) exceeds gpt-5.4
and gemini-3.1-pro but remains 2.1 points below claude-opus-4-8 on unrounded
accuracies. Channel B requires embedding access and was not run on the commercial
APIs, so these are contextual comparisons and not paired ones. LLaVA-OneVision
cannot reliably map the supplied directions to labels, and claude-opus-4-8, the
strongest baseline in the group, loses 6.9 points under Channel A. Evidence helps
only when a model can reconcile it with its own prior.

\paragraph{Cross-benchmark transfer.}
On the 120-item CineTechBench composition subset, zero-shot accuracy is 65.0\% and
Channel A reaches 64.2\%, a difference of one item. \emph{Central} rises from
78.6\% to 100\%, and \emph{Rule of Thirds} drops by 3.5 points because the leakage
guard suppresses answer-vocabulary wording. Transfer is neutral overall, and the
outcome depends on how the geometry maps onto a benchmark's terminology.

\subsection{Effect on Non-Composition Questions}
Appending composition evidence to the 3{,}093 questions in ShotBench's seven non-composition dimensions raises accuracy from 52.9\% to 56.4\%. No dimension drops by more than one point (camera angle, $-0.9$), while lens size rises by 10.6 and lighting by 6.4. Roughly 2--3 of the aggregate points come from the answer-format instruction, which reduces parse failures from 3.8\% to zero, and we do not attribute that part to geometry.

\section{Discussion}
The experiments show a trade-off between the model's own prior and the injected evidence. Measurements placed before the task change how the model reads the image and question, while post-question placement with gating preserves more correct baseline predictions and still gains on the directional classes.

\subsection{Evidence for the Integration Interpretation}
Four controls point to the interface rather than to the geometry alone. The rule
classifier and the evidence-free chain-of-thought control both stay below C0.
Rewording the block at the pre-question position also leaves accuracy below C0,
while moving the same facts after the question adds 7.1 points. Corruption affects
the naive and the calibrated interface differently with the image and decoder
fixed. Both channels consume the same vector and freeze the same VLM, yet Channel
A gates statements and Channel B learns four input embeddings, which weakens a
template-only explanation of the gain. Taken together, these controls locate the effect in the
interface: what changes across conditions is how much of the injected block the
model acts on.

We identify three error sources. \emph{Measurement error} comes from a wrong box or
pseudo-saliency map; mass corruption quantifies it. \emph{Integration error} means
a model over-follows a value or cannot map it to an option; prompt position and the
nine-model study expose it. \emph{Terminology error} leaves visible geometry
disconnected from a cinematic label, as with \emph{Short side}. These require,
respectively, better perception, trust calibration, and targeted language grounding.

The persistent \emph{Short side} failure separates visual evidence from language grounding. Every VLM, including the trained specialist, scores zero on this class under C0 and Channel A, while an F9 rule reaches 27.5\%. ShotQA contains 39 \emph{Short side} and 89 \emph{Symmetrical} samples against roughly 1{,}700 per frequent class. The geometry is therefore recoverable from the features while the cinematic label itself remains underlearned.

\subsection{Choosing Between the Two Interfaces}
The two channels share the extraction stack and the 24-dimensional vector, and they
occupy different deployment points. Channel A changes only the text sent to the
model, which suits closed APIs and settings where a human has to inspect the facts
that were exposed. Its cost is model dependence, since the gates control the
reliability of the measurements and not the willingness of the receiving model to
follow them. Channel B requires access to the input embeddings and improves the
same C0 model by 11.5 points while training 0.1\% of a 7B backbone. Added capacity
does not help by itself, since an eight-token, one-epoch variant reaches 60.1\%,
below the 61.0\% of the final setting.

A change of deployment distribution requires re-estimating the
training-distribution percentiles and re-checking the receiving model on that
distribution.

\subsection{Threats to Validity and Scope}\label{sec:validity}
\paragraph{Construct validity.}
Composition labels compress a subjective visual judgement into one option, and
exact-match accuracy cannot represent plausible secondary readings of an image.
Our nine features cover subject placement, balance, symmetry, depth, and prominent
lines, and they omit colour, motion, gaze, semantic emphasis, and temporal
editing. The zero accuracy on \emph{Short side} and the decline on
\emph{Balanced} mark that boundary.

\paragraph{Internal validity.}

All primary conditions use the same 479 items, official prompt, greedy decoder,
and paired transition analysis; experts and VLM weights stay frozen, gates use only
ShotQA training statistics, and test labels never enter feature construction. These
controls do not eliminate selection effects from design iteration. The
$3{\times}3$ threshold sweep and the fixed train/validation split show local
stability; we did not run nested model selection across independent datasets.
Detector failures are correlated, since a wrong dominant box changes centroid,
area, depth, and visual mass together, and our all-field corruption reproduces one
such failure family.

\paragraph{Statistical validity.}

The exact McNemar test applies because the systems answer the same questions, and
the primary Channel A--C0 comparison remains significant ($p{=}0.0018$). With 479
questions the power is limited for rare labels and small model differences.
Channel B exceeds Channel A by 4.8 points, but their paired comparison is
borderline ($p{=}0.058$). The per-class, threshold-sweep, and nine-model results
are therefore diagnostic, and we report transition counts alongside them.

\paragraph{External and deployment validity.}
CineTechBench is neutral overall, which shows that a taxonomy change can erase the
primary gain, and two of nine models regress. Neither scale nor a strong zero-shot
score guarantees that a new VLM will consult the block appropriately. Channel A gates individual lines but never
abstains, so an undetected and confidently wrong measurement can still change an
answer. A new camera domain,
language, or label set requires re-estimating evidence precision and validating
the receiving model.

\section{Conclusion}
On ShotBench composition, pre-question evidence lowers clean accuracy and follows
corrupted measurements. Post-question gating improves the frozen VLM by 6.7 points,
and a soft-token projector improves it by 11.5 points. Across nine models the same
text interface produces gains, negligible changes, and regressions. Explicit
geometry therefore helps when its interface is calibrated both for the reliability
of the measurements and for the receiving model.

\section*{Acknowledgments}
This work was supported by the Innovation Fund Project for Graduate Students of Bohai University.

\section*{Competing Interests}
The authors have no competing interests to declare that are relevant to the content of this article.

\bibliographystyle{plainnat}
\bibliography{refs}

@inproceedings{liu2025shotbench,
  title={ShotBench: Expert-Level Cinematic Understanding in Vision-Language Models},
  author={Liu, Hongbo and others},
  booktitle={NeurIPS},
  year={2025},
  note={arXiv:2506.21356}
}

@article{wu2025refineshot,
  title={RefineShot: Rethinking Cinematography Understanding with Foundational Skill Evaluation},
  author={Wu, Hang and others},
  journal={arXiv preprint arXiv:2510.02423},
  year={2025}
}

@article{wang2025cinetechbench,
  title={CineTechBench: A Benchmark for Cinematographic Technique Understanding and Generation},
  author = {Wang, Xinran and Xu, Songyu and Shan, Xiangxuan and others},
  journal={arXiv preprint arXiv:2505.15145},
  year={2025}
}

@inproceedings{rao2020movieshots,
  title={A Unified Framework for Shot Type Classification Based on Subject Centric Lens},
  author={Rao, Anyi and others},
  booktitle={ECCV},
  year={2020},
  note={arXiv:2008.03548}
}

@inproceedings{huang2020movienet,
  title={MovieNet: A Holistic Dataset for Movie Understanding},
  author={Huang, Qingqiu and others},
  booktitle={ECCV},
  year={2020},
  note={arXiv:2007.10937}
}

@article{wu2026camreasoner,
  title={CamReasoner: Reinforcing Camera Movement Understanding via Structured Spatial Reasoning},
  author={Wu, Hang and others},
  journal={arXiv preprint arXiv:2602.00181},
  year={2026}
}

@inproceedings{feng2026geometryguided,
  title={Geometry-Guided Camera Motion Understanding in VideoLLMs},
  author={Feng, Haoan and Musunuri, Sri Harsha and Su, Guan-Ming},
  booktitle={CVPR Workshops},
  pages={8423--8433},
  year={2026},
  note={arXiv:2603.13119}
}

@article{zhang2026camcue,
  title={Predicting Camera Pose from Perspective Descriptions for Spatial Reasoning},
  author={Zhang, Xuejun and others},
  journal={arXiv preprint arXiv:2602.06041},
  year={2026}
}

@inproceedings{you2026photoframer,
  title={PhotoFramer: Multi-modal Image Composition Instruction},
  author = {You, Zhiyuan and Wang, Ke and Zhang, He and others},
  booktitle={CVPR},
  pages={10197--10207},
  year={2026},
  note={arXiv:2512.00993}
}

@inproceedings{wei2018goodview,
  title={Good View Hunting: Learning Photo Composition from Dense View Pairs},
  author={Wei, Zijun and others},
  booktitle={CVPR},
  year={2018}
}

@inproceedings{li2020composing,
  title={Composing Good Shots by Exploiting Mutual Relations},
  author={Li, Debang and others},
  booktitle={CVPR},
  year={2020}
}

@article{bai2025qwen25vl,
  title={Qwen2.5-VL Technical Report},
  author={Bai, Shuai and others},
  journal={arXiv preprint arXiv:2502.13923},
  year={2025}
}

@article{zhu2025internvl3,
  title={InternVL3: Exploring Advanced Training and Test-Time Recipes for Open-Source Multimodal Models},
  author={Zhu, Jinguo and others},
  journal={arXiv preprint arXiv:2504.10479},
  year={2025}
}

@article{li2024llavaonevision,
  title={LLaVA-OneVision: Easy Visual Task Transfer},
  author={Li, Bo and others},
  journal={arXiv preprint arXiv:2408.03326},
  year={2024}
}

@inproceedings{liu2024groundingdino,
  title={Grounding DINO: Marrying DINO with Grounded Pre-Training for Open-Set Object Detection},
  author={Liu, Shilong and others},
  booktitle={ECCV},
  year={2024},
  note={arXiv:2303.05499}
}

@inproceedings{yang2024depthanythingv2,
  title={Depth Anything V2},
  author={Yang, Lihe and others},
  booktitle={NeurIPS},
  year={2024},
  note={arXiv:2406.09414}
}

@article{vongioi2010lsd,
  title={LSD: A Fast Line Segment Detector with a False Detection Control},
  author={von Gioi, Rafael Grompone and others},
  journal={IEEE TPAMI},
  volume={32},
  number={4},
  year={2010}
}

@article{wang2004ssim,
  title={Image Quality Assessment: From Error Visibility to Structural Similarity},
  author={Wang, Zhou and Bovik, Alan C. and Sheikh, Hamid R. and Simoncelli, Eero P.},
  journal={IEEE TIP},
  volume={13},
  number={4},
  year={2004}
}

@article{mcnemar1947,
  title={Note on the Sampling Error of the Difference Between Correlated Proportions or Percentages},
  author={McNemar, Quinn},
  journal={Psychometrika},
  volume={12},
  number={2},
  year={1947}
}

@inproceedings{johnson2017clevr,
  title={CLEVR: A Diagnostic Dataset for Compositional Language and Elementary Visual Reasoning},
  author = {Johnson, Justin and Hariharan, Bharath and van der Maaten, Laurens and others},
  booktitle={CVPR},
  pages={2901--2910},
  year={2017}
}

@inproceedings{hudson2019gqa,
  title={GQA: A New Dataset for Real-World Visual Reasoning and Compositional Question Answering},
  author={Hudson, Drew A. and Manning, Christopher D.},
  booktitle={CVPR},
  pages={6700--6709},
  year={2019}
}

@inproceedings{yang2019spatialsense,
  title={SpatialSense: An Adversarially Crowdsourced Benchmark for Spatial Relation Recognition},
  author={Yang, Kaiyu and Russakovsky, Olga and Deng, Jia},
  booktitle={ICCV},
  pages={2051--2060},
  year={2019}
}

@article{liu2023vsr,
  title={Visual Spatial Reasoning},
  author={Liu, Fangyu and Emerson, Guy and Collier, Nigel},
  journal={TACL},
  volume={11},
  pages={635--651},
  year={2023},
  doi={10.1162/tacl_a_00566}
}

@inproceedings{kamath2023whatsup,
  title={What's ``up'' with Vision-Language Models? Investigating Their Struggle with Spatial Reasoning},
  author={Kamath, Amita and Hessel, Jack and Chang, Kai-Wei},
  booktitle={EMNLP},
  pages={9161--9175},
  year={2023},
  doi={10.18653/v1/2023.emnlp-main.568}
}

@inproceedings{chen2024spatialvlm,
  title={SpatialVLM: Endowing Vision-Language Models with Spatial Reasoning Capabilities},
  author = {Chen, Boyuan and Xu, Zhuo and Kirmani, Sean and others},
  booktitle={CVPR},
  pages={14455--14465},
  year={2024}
}

@inproceedings{gupta2023visprog,
  title={Visual Programming: Compositional Visual Reasoning Without Training},
  author={Gupta, Tanmay and Kembhavi, Aniruddha},
  booktitle={CVPR},
  pages={14953--14962},
  year={2023}
}

@inproceedings{suris2023vipergpt,
  title={ViperGPT: Visual Inference via Python Execution for Reasoning},
  author={Sur{\'\i}s, D{\'\i}dac and Menon, Sachit and Vondrick, Carl},
  booktitle={ICCV},
  pages={11888--11898},
  year={2023}
}

@inproceedings{schick2023toolformer,
  title={Toolformer: Language Models Can Teach Themselves to Use Tools},
  author = {Schick, Timo and Dwivedi-Yu, Jane and Dessi, Roberto and others},
  booktitle={NeurIPS},
  volume={36},
  year={2023}
}

@inproceedings{alayrac2022flamingo,
  title={Flamingo: A Visual Language Model for Few-Shot Learning},
  author = {Alayrac, Jean-Baptiste and Donahue, Jeff and Luc, Pauline and others},
  booktitle={NeurIPS},
  volume={35},
  year={2022}
}

@inproceedings{li2023blip2,
  title={BLIP-2: Bootstrapping Language-Image Pre-training with Frozen Image Encoders and Large Language Models},
  author={Li, Junnan and Li, Dongxu and Savarese, Silvio and Hoi, Steven},
  booktitle={ICML},
  volume={202},
  pages={19730--19742},
  year={2023}
}

@inproceedings{lester2021prompt,
  title={The Power of Scale for Parameter-Efficient Prompt Tuning},
  author={Lester, Brian and Al-Rfou, Rami and Constant, Noah},
  booktitle={EMNLP},
  pages={3045--3059},
  year={2021},
  doi={10.18653/v1/2021.emnlp-main.243}
}

@inproceedings{li2021prefix,
  title={Prefix-Tuning: Optimizing Continuous Prompts for Generation},
  author={Li, Xiang Lisa and Liang, Percy},
  booktitle={ACL-IJCNLP},
  pages={4582--4597},
  year={2021},
  doi={10.18653/v1/2021.acl-long.353}
}

@inproceedings{jia2022vpt,
  title={Visual Prompt Tuning},
  author = {Jia, Menglin and Tang, Luming and Chen, Bor-Chun and others},
  booktitle={ECCV},
  pages={709--727},
  year={2022},
  doi={10.1007/978-3-031-19827-4_41}
}

@inproceedings{dorkenwald2024pin,
  title={PIN: Positional Insert Unlocks Object Localisation Abilities in VLMs},
  author={Dorkenwald, Michael and Barazani, Nimrod and Snoek, Cees G. M. and Asano, Yuki M.},
  booktitle={CVPR},
  pages={13548--13558},
  year={2024}
}

@inproceedings{geifman2019selectivenet,
  title={SelectiveNet: A Deep Neural Network with an Integrated Reject Option},
  author={Geifman, Yonatan and El-Yaniv, Ran},
  booktitle={ICML},
  volume={97},
  pages={2151--2159},
  year={2019}
}

@inproceedings{guo2017calibration,
  title={On Calibration of Modern Neural Networks},
  author={Guo, Chuan and Pleiss, Geoff and Sun, Yu and Weinberger, Kilian Q.},
  booktitle={ICML},
  volume={70},
  pages={1321--1330},
  year={2017}
}

@inproceedings{mai2016aesthetics,
  title={Composition-Preserving Deep Photo Aesthetics Assessment},
  author={Mai, Long and Jin, Hailin and Liu, Feng},
  booktitle={CVPR},
  pages={497--506},
  year={2016}
}

@article{liu2024lost,
  title={Lost in the Middle: How Language Models Use Long Contexts},
  author = {Liu, Nelson F. and Lin, Kevin and Hewitt, John and others},
  journal={TACL},
  volume={12},
  pages={157--173},
  year={2024},
  doi={10.1162/tacl_a_00638}
}

\end{document}